\documentclass[letterpaper, 10 pt, conference]{ieeeconf}  

\IEEEoverridecommandlockouts                              

\usepackage{amsthm}
\usepackage{subfiles}
\usepackage{wrapfig}
\usepackage{lipsum}
\usepackage{footnote}

\usepackage{booktabs}
\usepackage{algorithm}
\usepackage{algpseudocode}
\usepackage{algorithmicx}
\usepackage{subcaption}
\usepackage{graphicx}

\usepackage{hyperref}

\usepackage{amsmath}

\newcommand{\norm}[1]{\left\lVert#1\right\rVert}
\newtheoremstyle{named}{}{}{\itshape}{}{\bfseries}{.}{.5em}{\thmnote{#3}#1}
\theoremstyle{named}
\newtheorem*{namedtheorem}{}

\title{Impact of Heterogeneity in Multi-Robot Systems on Collective Behaviors Studied Using a Search and Rescue Problem}

\author{Sanjay Sarma O V$^{1}$, Ramviyas Parasuraman$^{2}$ and Ramana Pidaparti$^{3}$
\thanks{$^{1}$ School of Electrical and Computer Engineering, College of Engineering, University of Georgia, Athens, GA 30602, USA
        {\tt\small sanjaysarmaov@uga.edu}}%
\thanks{$^{2}$HeRo Lab, Department of Computer Science, The University of Georgia, Athens, GA 30602, USA 
        {\tt\small ramviyas@uga.edu}}%
\thanks{$^{3}$ School of Environmental, Civil, Agricultural and Mechanical Engineering, College of Engineering, The University of Georgia, Athens, GA 30602, USA 
        {\tt\small rmparti@uga.edu}}%
}

\begin{document}
\maketitle

\begin{abstract}
Many species in nature demonstrate symbiotic relationships leading to emergent behaviors through cooperation, which are sometimes beyond the scope of the partnerships within the same species. These symbiotic relationships are classified as mutualism, commensalism, and parasitism based on the benefit levels involved. While these partnerships are ubiquitous in nature, it is imperative to understand the benefits of collective behaviors in designing heterogeneous multi-robot systems (HMRS). In this paper, we investigate the impact of heterogeneity on the performance of HMRS applied to a search and rescue problem. The groups consisting of searchers and rescuers, varied in the individual robot behaviors with multiple degrees of functionality overlap and group compositions, demonstrating various levels of heterogeneity. We propose a new technique to measure heterogeneity in the agents through the use of Behavior Trees and use it to obtain heterogeneity informatics from our Monte Carlo simulations. The results show a positive correlation between the groups’ heterogeneity measure and the rescue efficiency demonstrating benefits in most of the scenarios. However, we also see cases where heterogeneity may hamper the group’s abilities pointing to the need for determining the optimal heterogeneity in group required to maximally benefit from HMRS in real-world applications.
\end{abstract}
\section{INTRODUCTION}

Disasters cause severe disruption of systems impacting humans, materials, the environment, and the economy. Many times, efficiently responding and reacting to these disasters may get hard for humans, which mostly involves conducting search and rescue operations involving other humans or material \cite{cubber2017introduction}. In such caes, use of multi-UAV, UGVs, and UUVs systems have proven to be beneficial \cite{erdelj2017help}. Specifically, we emphasize the importance of Heterogeneous Multi-Robot Systems (HMRS) in urban search and rescue (USAR) applications \cite{liu2016multirobot,polka2017use,yang2020needs}.
HMRS include robots of different types with structural and functional differences that are similar to animals from different species. 

In natural living systems, partnerships between different animal species are ubiquitous in nature, which help them in their survival, betterment, and evolution. These partnerships are broadly classified as mutualism, commensalism, and parasitism, based on the type of benefit or harm a participating species has from the other.  In a mutualistic type of relationship, both the species are benefited through a partnership \cite{10.2307/3036036,Thompson372}, in commensalism, one is benefited while the other is not harmed \cite{dales1957interrelations}, and in parasitism, one species is harmed while the other one benefits \cite{wobeser2008parasitism}.

Heterogeneity in a group can either arise dynamically due to physical constraints or can be a macro property in a group. For example, a mostly homogeneous robot group can be heterogeneous at a micro level, with minor differences in sensor or actuation levels. On the other hand, a macro level heterogeneity can be due to various types of robots (UAVs and UGVs) within a group. Twu et al., \cite{twu2014measure} define heterogeneity in a multi-agent system as a product of complexity and disparity, where complexity refers to the variety in the group and disparity refers to the distinction between the agents within a group.

Over the past decade, various HMRS strategies have been developed for collective path planning, exploration, self-organization, formation control, and disaster management. Rizk et al. present a comprehensive survey on the existing state of the art cooperative HMRS \cite{rizk2019cooperative}, along with the limitations and challenges faced in this domain. 
Recent research in the use of HMRS for USAR is focused on developing efficient algorithms and strategies in decision making, development of novel networks, reducing human effort at low-level control, etc. \cite{pippin2011bayesian,7440563}.
Predominantly these strategies are designed for the collective accomplishment of a task through cooperation \cite{dadvar2020multiagent}, in which a primary task (a mission) is decomposed into sub-tasks and assigned to robots through techniques like performance assessment and auction \cite{darintsev2019methods, hunt2014consensus}. These behaviors are either decided upfront or dynamically changed with time and circumstances in a mission,  based on the performance at the mission level.

A similarity between the HMRS strategies can also be drawn to the dynamic nature of symbiotic relationships in nature, where a group or species varies its relationships with others based on the benefit levels it perceives at a given point of time. Drawing inspiration from these dynamics in symbiotic relationships, we deem that an understanding of the need for heterogeneity and functional overlap between agents is crucial for designing an optimal team composition and task allocation to successfully deploy HMRS in real-world USAR applications \cite{liu2013robotic}. 

Therefore, in this paper, we present an analysis of the effect of heterogeneity and functionality overlap on an HMRS applied to a Search and Rescue problem.
Specifically, we design an USAR problem with two types of agents, searchers and rescuers, where searchers lookout for a targets and rescuers retrieve them. 
While these robots are distinct in primary functionalities, we vary the amount of functionality overlap between the agents to capture their complex relationships.

We build upon the work in \cite{twu2014measure} and  propose a new exemplar technique to measure the heterogeneity (functionality overlap) in robots and multi-agent systems by exploiting a distance function applied on the state-action plan of the robots represented through Behavior Trees \cite{colledanchise2018behavior,colledanchise2018learning}. We create Monte Carlo simulations of several mixture of robot groups with varying degree of heterogeneity and diversity. Through the results, we observe the system-level global SAR performance in terms of cost and efficiency to analyze and discuss the impact of heterogeneity in such applications. 

Our analysis demonstrate the extent at which this functionality overlap determines the performance of the system. We present the formulation of the USAR problem in Sec. II, followed by our proposed approach in Sec. III to measure heterogeneity and conducting simulations on a Unity game engine. 
We present the results obtained from various experiments conducted on different agent combinations in Sec. IV and finally conclude the paper with arguments on the benefits of heterogeneity and its shortcomings in Sec. V.

\section{SAR Problem Formulation}
We investigate the need for heterogeneity and functional overlap between agents in HMRS through a Search and Rescue problem by introducing two types of robots called searchers and rescuers, whose primary task involves the retrieval of targets scattered in a configuration space. This section presents the definitions for the search space, sensor and control models used. \par

\subsection{Search Space}
The search area is a rectangular configuration space given by $A=[0,x]\times[0,y]$. The target points $T$ (e.g, victims in USAR) are blocks or points which, are randomly located in the space and should be moved to one of the collection or retrieval points in $C$. The number of target points (we use target points and treasure blocks interchangeably), scattered in the configuration space is $n_t$ and $n_r,n_s$ and $n_c$ correspond to the number of rescuers, searchers and collection points in the configuration space respectively. Further, $n_H=n_r+n_s$ is the total number of heterogeneous agents and $n_{h_a}$ are the number of acceptable hosts, and $h_a \subseteq H$, which are changed with the scenarios (defined in Sec.~\ref{sec:RescStrat}).

The location of the target is given by, $T_i \in A$ and $ T=[T_1,T_2\dots T{n_t}]$. Similarly, the collection points $C$ are fixed and given by, $C_i \in A$ and $C=[C_1,C_2 \dots C_{n_c}]$. The target points and collection points are static through the simulations and hence, $\dot{T}=0$ and $\dot{C}=0$.\par

\subsection{Sensor Models}

\begin{figure}[t]
\centering
\includegraphics[width=8cm]{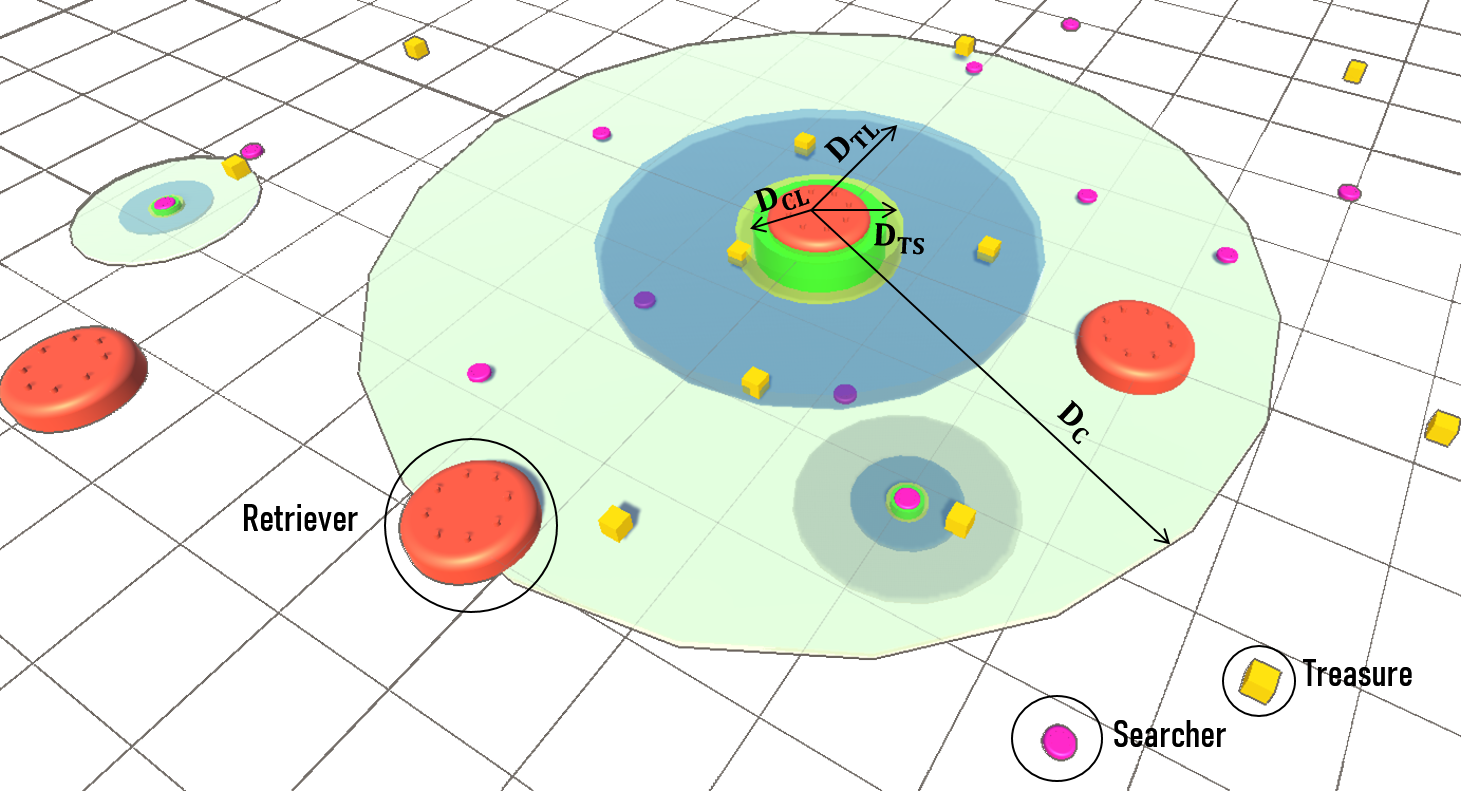}
\setlength{\belowcaptionskip}{-10pt}
\caption{The ranges of communication,collision detector, short range target detector, long range target detector are labeled as $D_C,D_{CL},D_{TS},D_{TL}$,  respectively. }

\label{pic:SensorPic}
\end{figure}

Each of the searchers and rescuers have a suite of sensors for detecting collisions, for short and long range detection of target and communication, whose ranges are $D_{CL},D_{TS},D_{TL}$ and $D_C$ respectively. Also, the ranges of these sensors follow the order, $D_{CL} < D_{TS} < D_{TL} < D_C$ and the detection ranges associated with each of the sensors is shown in Fig.~\ref{pic:SensorPic}. 

\subsubsection{Collision Detector}

The first type of sensor is for collision detection with other agents and the walls. Its field of collision at any given point of time is
\begin{equation}
F_{{CL}_i}=F_{CL}(H_i,D_{CL})
\end{equation}
This represents a disc of detection with radius $D_{CL}$ around agent $H_i$. We model this sensor to also estimate the probability of collision $P_{CL}$ and the point of collision $V_{CL}$ by the following equations
\begin{equation}
    P_{CL_{ji}}=P_{CL}(H_j,H_i)=\begin{cases}
    0 : d_{CL_i}> D_{CL}\\
    1 : d_{CL_i} \le D_{CL}
    \end{cases}
\end{equation}
\vspace{-4mm}
\begin{equation}
    V_{CL_{ji}} =V_{CL}(H_j,H_i) \\
     =  \begin{cases}
    H_j-H_i:&d_{CL_i} \le D_{CL} \\& \lor P_{CL_j}=1 \\
    0 : P_{CL_j}=0
            \end{cases}
\end{equation}
where $d_{CL_i}=\norm{H_j-H_i}$ and $j \in [1,n_H]-\{i\}, i \in [1,n_H]$. When multiple collisions are detected simultaneously, a resultant of all the $V_{CL}$ is generated by the collision detector, with respect to the robot.
\begin{equation}
    V_{CL_i}=\sum_{j=1}^{n_H-\{i\}} V_{CL_j}
\end{equation}

\par

\subsubsection{Short Range Target Detector}
This sensor was modeled to detect only the presence of a target point with in its range, and cannot generate any relative position vectors. The detection disc radius is $D_{TS}$ and the field of view for the sensor is defined as,
\begin{equation}
    F_{TS_i}=F_{TS}(H_i,D_{TS})
\end{equation}
its probability of detection is given by,
\begin{equation}
   P_{TS_i}=P_{TS}(T,H_i)=\begin{cases}
    0:d_{TS_i}>D_{TS}\\
    1:d_{TS_i}\leq D_{TS}
    \end{cases}
\end{equation}
where, $d_{TS_i}=\norm{T_j-H_i}$ and $1\le j \le n_t,i \in [1,n_H]$

\subsubsection{Long Range Target Detector} 
This sensor is similar to a short range target detector with additional ability to detect all the target points in its proximity and also generate a position vector to the nearest one, within a disc of radius $D_{TL}$. The field of view of the sensor, probability of detection and position vector are
\begin{equation}
    F_{TL_i}  =   F_{TL}(h_{a_i},D_{TL}) 
\end{equation}
\vspace{-8mm}
\begin{equation}
\begin{split}
P_{TL_i}  =P_{TL}(T,h_{a_i})  = & \begin{cases}
    \begin{split}
        0:& d_{TL_i}>D_{TL} \lor P_{TS_i}=1\\
    \end{split}\\
    \begin{split}
        1:d_{TL_i}\le D_{TL} \land P_{TS_i}=0
    \end{split}
    \end{cases}\\ &,\forall j\in [1,n_{h_a}]-\{i\} 
    \end{split} 
\end{equation}
\vspace{-8mm}
\begin{equation}
\begin{split}
\label{eq:vt}
    V_{T_i}=V_T(T,h_{a_i})  =  \begin{cases}
    \begin{split}
     T_{best}-&h_{a_i} \mid T_{best} \in T \\&\land \norm{T_{best}-h_{a_i}}= d_{tmin} \\&:P_{TL_i}=1
    \end{split}\\
   0: P_{TL_i}=0
    \end{cases}\\
  d_{tmin} = \min\limits_{\forall j \in n_t}\norm{T_j -h_{a_i}} ,
    \end{split} 
\end{equation}
where $d_{TL}=\norm{T_j-h_{a_i}}$ and $1\le j \le n_t$.

\subsubsection{Communicator}

A communicator communicates with the nearest active transmitter in its range and its field of view is defined as,
\begin{equation}
    F_{C_i} =F_c(h_{a_i},D_c)
\end{equation}
Here, we maintain that the probability of communication is based on the state of the other agent under interaction. Only, those agents which have a target detected by its long-range sensor is considered as an active transmitter. Inversely, this can also be stated as, an agent generates beacon on encountering a target by its long-range detector. The probability of detection and the position vector of the nearest active transmitter is given as,
\begin{equation}
\begin{split}
    P_{C_i}=P_C(h_a,h_{a_i})= \begin{cases}
    \begin{split}
        0:&\norm{h_{a_j} - h_{a_i}}>D_c \\ &\lor P_{TL}(T,h_{a_j})=0 \\ &\lor  P_{TL_i}=1\\
    \end{split}\\
    \begin{split}
        1:&\norm{h_{a_j} - h_{a_i}}\le D_c \\ &\land P_{TL}(T,h_{a_j})=1 \\ & \land P_{TL_i}  =0
    \end{split}
    \end{cases} \\ ,\forall j\in [1,n_{h_a}]-\{i\}
    \end{split}
\end{equation}
\vspace{-4mm}
\begin{equation}
\begin{split}
\label{eq:vc}
    V_{C_i}=V_C&(h_a,h_{a_i})=\begin{cases}
    \begin{split}
    h&_{a_{best}}-h_{a_i}  \mid h_{a_{best}}\in h_a\\ & \land \norm{h_{a_{best}}-h_{a_i}}= d_{hmin} \\ &:P_C=1
    \end{split}\\
    0 : P_C=0
    \end{cases}\\
    &d_{hmin}=\min\limits_{\forall j \in [1,n_{h_a}]-\{i\}} \norm{h_{a_j} -h_{a_i}}
    \end{split}
\end{equation}

The minimization in Eq.~\eqref{eq:vc} corresponds to the selection of the nearest communicating agent. Hence, $V_C$ points to the nearest communicator.

\subsection{Motion Controller}
The robot movements are decided by a controller based on the following velocity equation, obtained by combining the position vectors from all the sensors and for a robot i, the resultant control vector $V_{con}$ is computed as,
\begin{equation}
    V_{con_i}=(V_{T_i}+V_{C_i}+V_{P_i}+V_{R_i})(1-P_{CL_i})-V_{CL_i} 
    \label{eq:control}
\end{equation}
Considering the velocity limits on an the agent, we compute a unit vector along $V_{con_i}$ and multiply it with the RobotMaxSpeed scalar as follows,
\begin{equation}
    v_{con_i}=\hat{V}_{con_i}\times RobotMaxSpeed
\end{equation}
Further, in Eq.~\eqref{eq:control}, $V_{R_i}$ is a random walk vector of $i^{th}$ agent, given by
\begin{equation}
    V_{R_i}=V_R(H_i)=\begin{cases}
    \begin{split}
        [v_x,v_y]\in R \mid & v_x,v_y \in [-1000,1000]\\ :& P_{TS_i} =0\\ &\land P_{TL_i}=0 \\&\land P_{TC_i}=0\\ &\land P_{CL_i} =0
    \end{split}\\
    \begin{split}
    0: P_{TS_i}=1 &\lor P_{TL_i}=1 \\ &\lor P_{C_i}=1 \lor P_{CL_i}=1
    \end{split}
    \end{cases}
\end{equation}

Retrieval vector $V_{P_i}$ points to the nearest collection point after the target is retrieved,
A target point is picked up when, $P_{TS}=1$. i.e., when a target falls within the short range target detector's vicinity and the agent has not picked up any target. This can be defined by a probability $P_p$, given by 
\begin{equation}
    P_p=\begin{cases}
    1: P_{TS}=1 \land  \text{no target on board}\\
    0: P_{TS}=0 \lor \text{target on board}
    \end{cases}
\end{equation}

Further, we define a collection point as a square region in the configuration space and retrieval vector $V_P$ is defined as
\begin{equation}
\label{eq:vp}
   V_{P_i}= V_P(C,H_i)=\begin{cases}
    \begin{split}
        C_{best}-H_i &\mid  C_{best} \in C\\ & \land \norm{C_{best}-H_i}=d_{cmin} \\ &:P_p=1 \\
    0: P_p=0
     \end{split}
    \end{cases} 
\end{equation}
where, $d_{cmin}=\min\limits_{\forall j \in [1,n_{c}]} \norm{C_j -H_i}$ 

The minimization in Eq.~\eqref{eq:vp} ensures that $V_P$ points to the nearest collection point.

\subsection{Heterogeneous Multi-Robot System}
The heterogeneous group in our study is a mix of searchers and rescuers. In this section, we define the searcher and rescuer behaviors and introduce the differences between them to establish the heterogeneity in the group. The sensor models defined in the previous section are common to both searchers and rescuers making them structurally similar. However, we maintain the distinction in controller responses making them functionally different.\par
\subsubsection{Searchers}
Searcher robots are the simplest type of agents in the current study. A searcher exists in either of the five states depending on which velocity term of the Eq.~\eqref{eq:control} is active. A searcher in the absence of any response from its sensors remains in a random walk state (state - 0). A searcher starts moving towards a target once detected by its long range sensor switching to state - 1 or move towards another agent upon receiving a beacon signal about a target from another agent in state - 2. A searcher switches to state - 3, when a target is detected by its short range sensor, stops nearby ($V_P=0$) and transmits a beacon signal. \par
\subsubsection{Rescuers}
\begin{figure*}
\centering
\includegraphics[width=16cm]{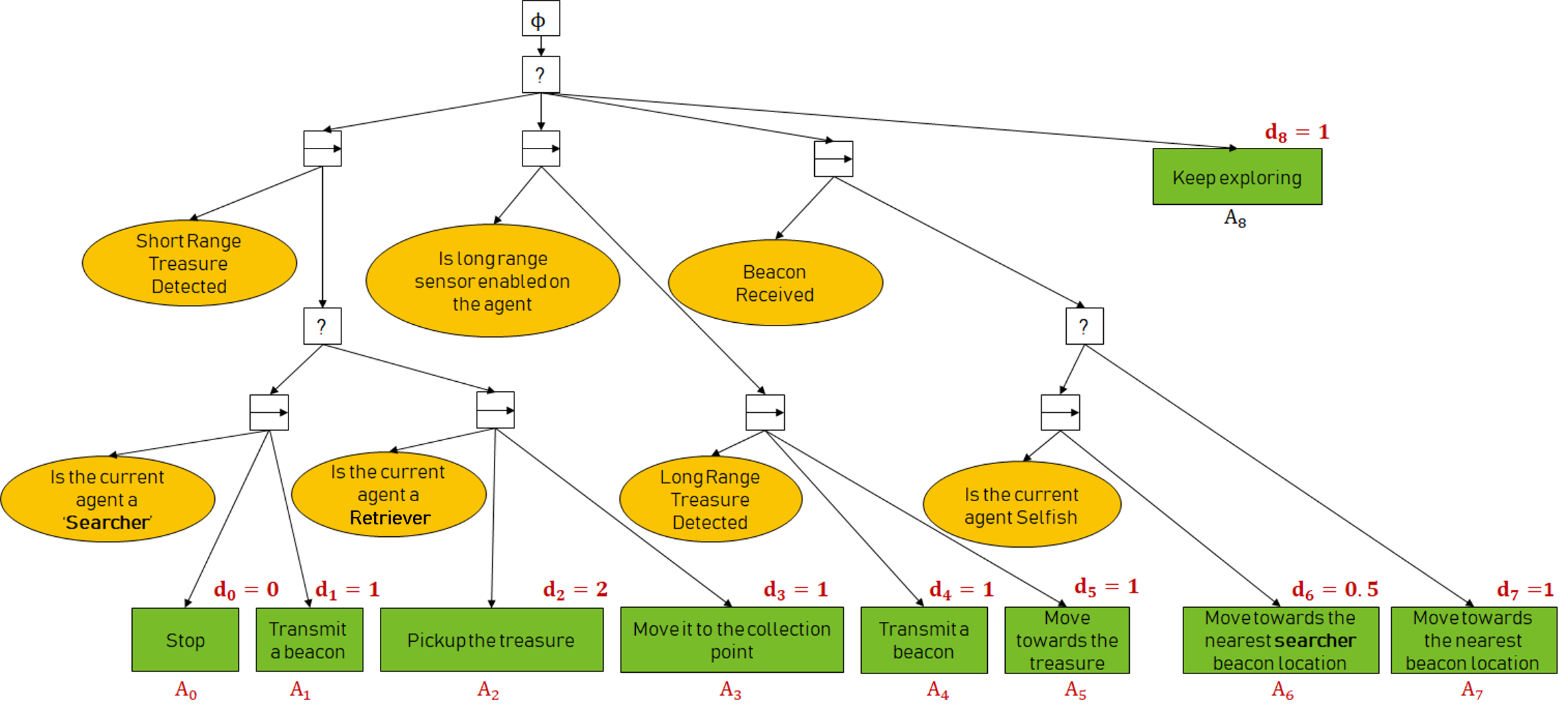}

\caption{Behavior tree (BT) summarising searcher and generic rescuer models. The distance values for every BT node ($d_0,...d_n$) are used to calculate the inter-species distance in Eq.~\ref{eq:Qmeasure} and Heterogeneity in the HMRS.}
\label{pic:SearcherBehave}
\vspace{-5mm}
\end{figure*}

Rescuers, on the other hand are similar to searchers, except for that they can pick up the target and can move it to one of the nearest collection points or otherwise $V_P\neq0$. Further, we maintain a distinction in the behavior of rescuers by modifying the rescuer abilities called strategies for our study. We present the generic rescuer and searcher behaviors through Behavior Tree representation \cite{colledanchise2018behavior} in Fig.~\ref{pic:SearcherBehave}, and pseudo-code for the controller in Alg.~\ref{algo:1}.


\begin{algorithm}
\caption{Agent Behavior}\label{algo:1}
\newcommand{\vars}{\texttt}
\newcommand{\func}{\textrm}

\begin{algorithmic}[1]

\Function{AgentControl}{$AgentType,AgentSpeed$}
\State $V_{CL},P_{CL} \leftarrow \func{CollisionSensor()}$

\State $P_{TS} \leftarrow \func{TargetShortRange()}$
\State $V_{TL},P_{TL} \leftarrow \func{TargetLongRange()}$
\State $V_{C},P_{C} \leftarrow \func{Communication()}$

\State $V_{R}\leftarrow \func{RandomWalk()}$
\If{$AgentType$ is \textit{'Rescuer'}}
\State $V_{P},P_{P} = \func{Retrieve()}$
\ElsIf{$AgentType$ is \textit{'Searcher'}}
\State $V_{P}\leftarrow 0$
\State $P_{P}\leftarrow1$
\EndIf
\State $V_{con} \leftarrow (V_{TL}+V_C+V_R+V_{P})\times (1-P_{CL})-V_{CL}$

\Return $v_{con}  \leftarrow \frac{V_{CTRL}}{|V_{CTRL}|}$

\EndFunction
\end{algorithmic}
\emph{Note: Sensor outputs and actuator commands are represented as functions with their corresponding names.}
\end{algorithm}

Further, it can be noted that a Long-range detector points a vector to the nearest target for both searchers and rescuer when the short-range one detects no target. This is in contrast to a short-range detector, which does not generate a vector; however, it flags the presence of treasure in the vicinity of the robot (detection range) for pickup in case of rescuers, stop, and transmit in case of searchers. This order of preference is according to the behavior tree presented in Fig. \ref{pic:SearcherBehave}.

\subsection{Rescuer Strategies}
\label{sec:RescStrat}
The models presented so far, are for generic rescuers. However, for the current study, we slightly modify the behavior of the rescuers to act selfish or blind. The selfishness factor is decided by the accessible agents set $h_a$. We model the strategies for rescuers as follows. It must be noted that, all the rescuers demonstrate some amount of searching abilities also, demonstrating functional overlap between the agents. Here $S$ and $R$ are the sets of searchers and rescuers, respectively.
\begin{namedtheorem}[Strategy 1]
All agents (S, R) have all sensors enabled and,
\begin{equation}
\forall S \in H \land R \in H, h_a= H
\end{equation}
\end{namedtheorem}
In Strategy 1, the rescuers are as good as searchers in searching targets and also communicate among themselves.
\begin{namedtheorem}[Strategy 2]
All agents (S, R) have all sensors enabled and,
\begin{eqnarray}
\label{eq:selfish1}
\forall S \in H, h_a= H ,
 \\
\forall R \in H, h_a= H-R = S 
\label{eq:selfish2}
\end{eqnarray}
\end{namedtheorem}
Here, the $h_a$ sets for searchers and rescuers are exclusive. Which means, while all the searchers communicate with other searchers and rescuers, rescuers on the other hand are selfish and listen only to other searchers. Here, all the sensors on the rescuers are enabled.
\begin{namedtheorem}[Strategy 3]
All the searchers have all the sensors enabled, for rescuers in addition to having $h_a$ defined by Eqs.~\eqref{eq:selfish1} and \eqref{eq:selfish2}, the long range target sensors are also disabled i.e., $P_{TL}=0$ always.
\end{namedtheorem}
In Strategy 3, the rescuers are blind to long range targets and also are selfish like in Strategy 2.
\section{Proposed Approach}
\subsection{Measure of Heterogeneity}
\label{sec:measure}
For our study on the effect of heterogeneity, we segregated the rescuer and searcher combinations into three different groups called scenarios. In the first scenario, we considered a pure homogeneous system with only rescuers and their population was varied from a high to a low value. The second scenario was with a constant number of rescuers but with increasing searchers starting from zero. Finally, in the third scenario, we maintained the total population of searchers and rescuers constant ($n_r+n_s=n_H=constant$), and varied the ratio of composition between them. Also, rescuers with different strategy types introduced in the previous section were tested in all three scenarios. A summary of these agent combinations is presented in Table \ref{tab:simHyper}.\par
\begin{table}[h]
\caption{Simulation Hyperparameters}
\label{tab:simHyper}
\begin{tabular}{clc}
\toprule
\textbf{Scenario} & \textbf{Agent Combination}  \\
\midrule
1        & $n_r = 5:5:50$, $n_s=0$ (Homogeneous)              \\
2        & $n_r= 25,n_s=0:5:50$                  \\
3        & $n_r+n_s=n_{h_a}=50,  n_s=0:5:45$      \\     
\midrule
\multicolumn{1}{r}{Target Points}&250\\
\multicolumn{1}{r}{Collection Points}&4\\
\multicolumn{1}{r}{Rescuer Strategies}&1,2,3\\
\multicolumn{1}{r}{Trials} &10\\
\multicolumn{1}{r}{Trial Duration} & 300 s @ 0.02s/ Iteration (Frame)\\
\bottomrule
\end{tabular}
\end{table}
We computed a heterogeneity measure for all the scenarios and strategies obtained from Twu et al. \cite{twu2014measure} work, in which they quantified heterogeneity in multi-agent systems as a product of complexity and disparity. Their complexity measure corresponds to group's entropy and disparity is given by Rao's quadratic entropy \cite{rao1982diversity} based on inter-species distance. In these lines, we compute the entropy of the current search and rescuer groups as,
\begin{equation}
    E(p_H)=-(p_r\log{p_r} +p_s\log{p_s})
\end{equation}
here, $p_r =\frac{n_r}{n_H}$ and $p_s =\frac{n_s}{n_H}$.\\
Rao's quadratic entropy is defined as,
\begin{equation}
    Q(p_H)=2p_rp_sd_{rs}^2
    \label{eq:Qmeasure}
\end{equation}
And, heterogeneity measure as,
\begin{equation}
    H(p_H)=E(p_H)Q(p_H)
\end{equation}\par
Here, $d_{rs}$ is the distance between the two species types (searchers and rescuers). For computing the distance, we assigned a score to each of the robot actions represented in the form of a behavior tree as shown in Fig. ~ \ref{pic:SearcherBehave}. The score assignment was based on the complexity of the task involved. For example, a target retrieval task is more complex than random walk in the configuration space. Also, for a robot receiving a beacon, a selfish robot masks the other rescuers and hence has lesser number of robots in its list to compute the minimum distance ($d_6=0.5<d_7=1$). We added the scores of all possible actions of the robots in a strategy and computed its ratio to the total maximum score (here its 8.5). 

\begin{figure}
\centering
\begin{subfigure}{0.98\linewidth}
\centering
\includegraphics[width=8.5cm]{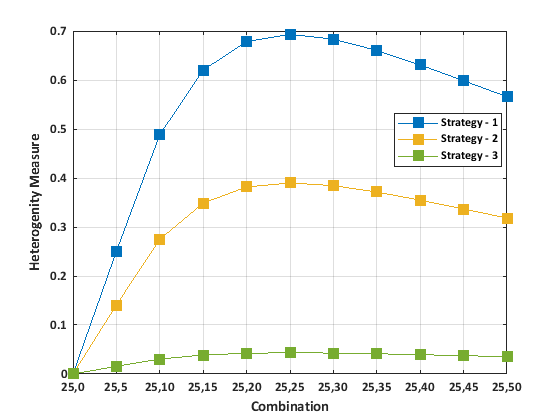}
\caption{Heterogeneity in Scenario 2 vs. \{R:S\} ratio. }
\label{pic:heteroMeasurea}
\end{subfigure}
\begin{subfigure}{0.98\linewidth}
\centering
\includegraphics[width=8.5cm]{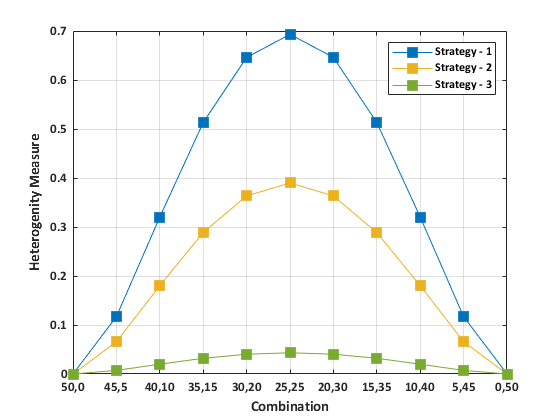}
\setlength{\belowcaptionskip}{-2pt}
\setlength{\abovecaptionskip}{-1pt}
\caption{Heterogeneity in Scenario 3 vs. \{R:S\} ratio.}
\label{pic:heteroMeasureb}
\end{subfigure}
\caption{Heterogeneity Measures for Strategies 1, 2 and 3. (x- axis denotes Rescuer and Searcher combinations - R:S).}
\label{pic:heteroMeasure}
\vspace{-5mm}
\end{figure}

For example, a searcher across all the strategies has no change in its behavior and hence the possible actions are ${A_0,A_1,A_4,A_5,A_7,A_8}$, which corresponds to a total score of 5. On the other hand, a rescuer of Strategy 1 has a possible action set of $A_2,A_3,A_4,A_5,A_7,A_8$ and a score of 7. Similarly for rescuers of strategies 2 and 3, the scores are 6.5 and 5.5 respectively. The heterogeneity measures computed for all the rescuer strategies in the scenarios 2 and 3 is presented in Fig. ~ \ref{pic:heteroMeasure}. It must be noted that, Scenario 1 has a homogeneous group and hence we did not present the heterogeneity measure graph.

Also, Fig. \ref{pic:heteroMeasure} shows that the heterogeneity peaks for combinations with an equal number of searchers and rescuers but drops as one agent type dominates the group. 

\subsection{Simulation Setup}
\begin{figure}[t]
\centering
\includegraphics[width=7.5cm]{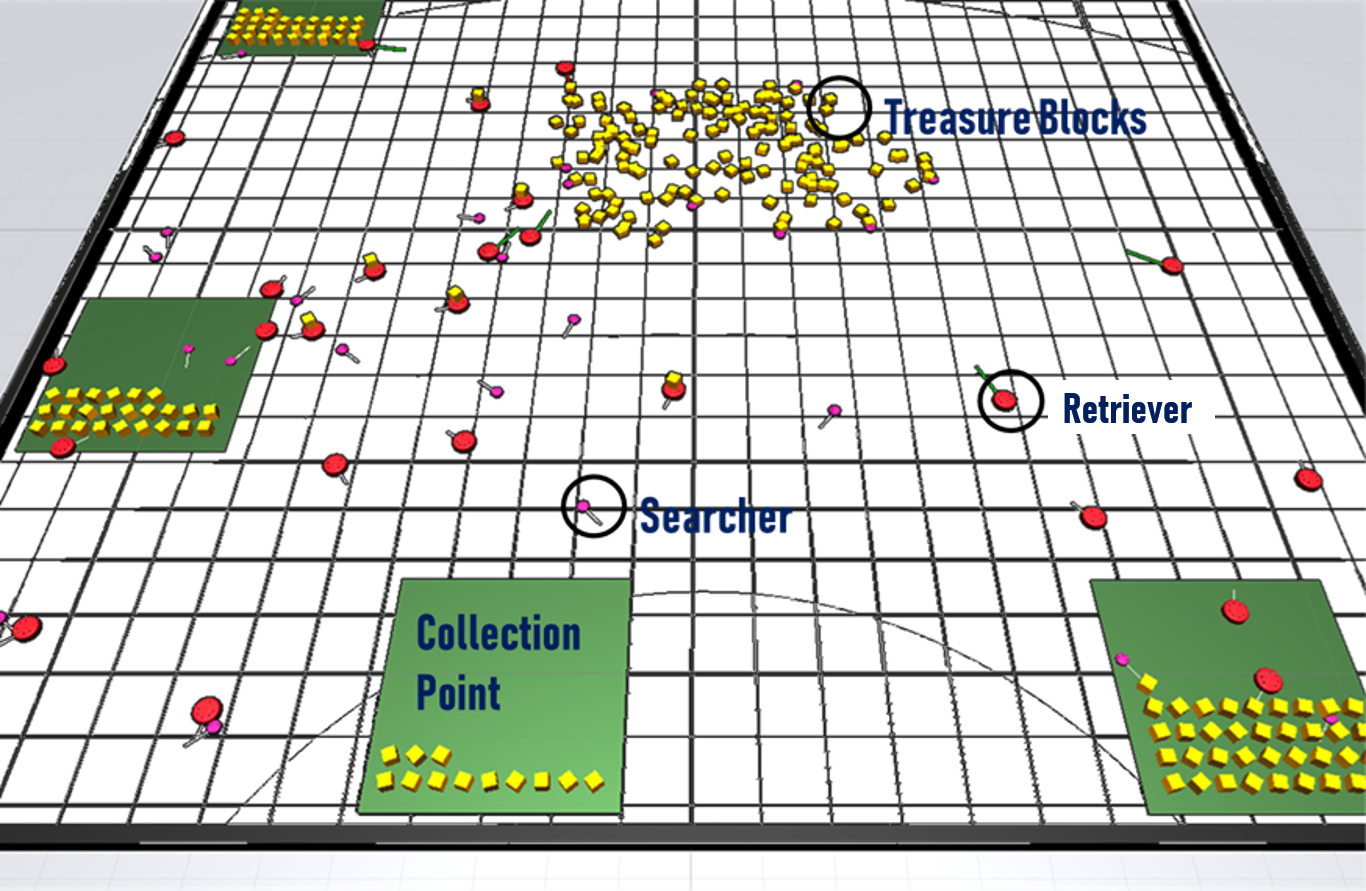}
\setlength{\belowcaptionskip}{-10pt}
\caption{The designed configuration space has four collection points (green), with targets (yellow) randomly placed. Searchers and rescuers are colored magenta and red respectively. A ray projecting from the agents indicates the direction of heading and their color represents the agent's state (Red - Long range target, Green - Communication, White - Collection or Random Walk.}
\label{pic:ConfigSpace}
\end{figure}

We developed a simulator in Unity game engine for studying the current SAR problem. The robots were modeled as game objects and the sensors were designed as cylindrical Colliders. An intermittent State Manager game object manages the sensor data and their states and communicates with a modelled controller. The final velocity vector $v_{controller}$ computed by the controller is sent to the actuator game object, which executes the lateral movements and target retrieval actions on the robots.
We designed the configuration space to have four collection points and their locations were predefined. Also, the targets were placed randomly for each trial. A snapshot of the configuration space designed for the simulations is presented in Fig.  \ref{pic:ConfigSpace}. 

\par

\subsection{Experiment Settings}
We conducted experiments on all strategy and scenario combinations and each combination 10 trials were run.
In each simulation trial, the target points and the searchers were randomly initialized in the configuration space and rescuers on the other hand were initialized in a zone near the right top corner of the configuration space in Fig.  \ref{pic:ConfigSpace}. Further, we maintained that the ten different target configurations (for 10 different trials) were the same across all the scenarios and strategy combinations for uniformity in analysis. This maintains that, the target placement changes across trials but remains constant across different scenarios. A total of 250 target points were placed in the configuration space across all the trials. A summary of the simulation hyperparameters is presented in table \ref{tab:simHyper}. A game manager was designed to run the simulations for all the hyperparameters read from a text file and it also logged the simulation data in a .csv file. 
A video demonstration of the experiments is available in at \url{http://hero.uga.edu/research/heterogeneity/}

\par 
\section{Results and Discussion}
\label{sec:ResD}

\begin{figure}[h]
\centering
\includegraphics[width=8.5cm]{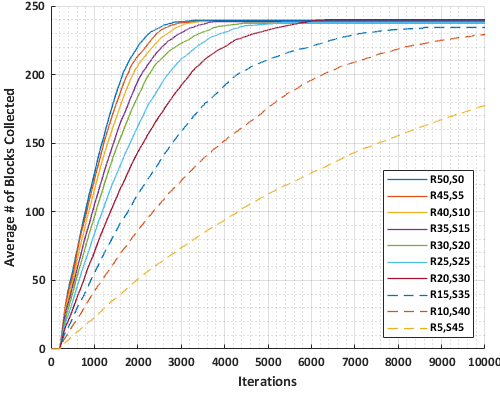}
\caption{Graph shows the average number of targets placed at the retrieval points over iterations, for rescuer strategy 1 and scenario 3. x scale limited to 10000 iterations.}
\label{pic:TimeSeries1}
\vspace{-4mm}
\end{figure}

\begin{figure}[t]

\centering
\begin{subfigure}{.47\linewidth}
\centering
\includegraphics[width=4cm]{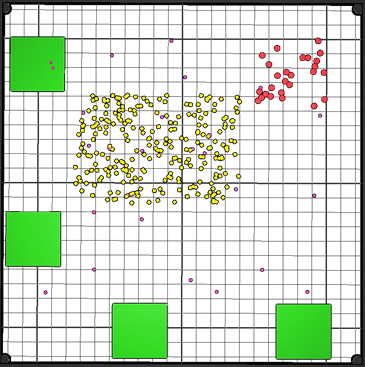}
\caption{Iteration 1}

\end{subfigure}
\begin{subfigure}{.47\linewidth}
\centering
\includegraphics[width=4cm]{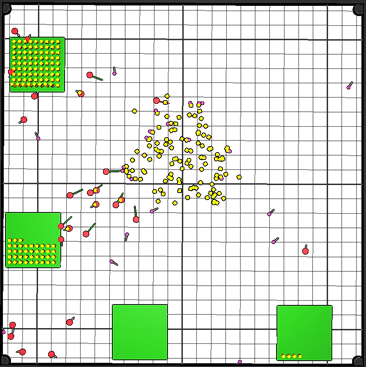}
\caption{Iteration 2000}

\end{subfigure}
\par\bigskip
    
\begin{subfigure}{.47\linewidth}
\centering
\includegraphics[width=4cm]{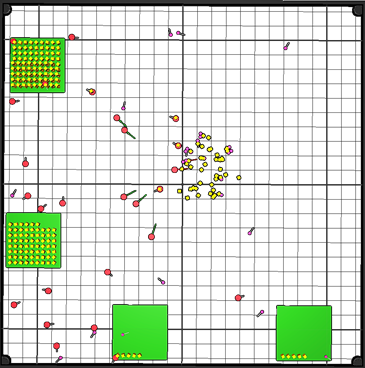}
\caption{Iteration 4000}

\end{subfigure}
\begin{subfigure}{.47\linewidth}
\centering
\includegraphics[width=4cm]{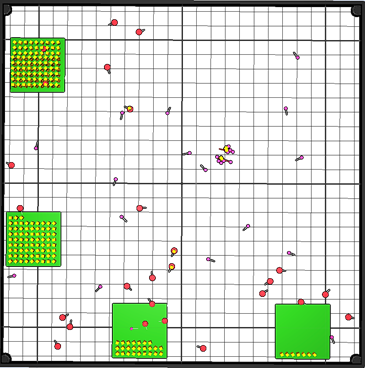}

\caption{Iteration 6000}

\end{subfigure}

\setlength{\belowcaptionskip}{-15pt}
\caption{Snapshots of the simulation at different intervals for 25 rescuers (Strategy 1) and 25 Searchers. }
\label{fig:snapShots}
\end{figure}

The data recorded for all the trials across the scenario and strategy combinations primarily consisted of the number of targets retrieved. The data collected across all the trials was averaged for each of the scenario and strategy combinations. Sample graphs for the average number of targets retrieved over time is presented in Fig.  \ref{pic:TimeSeries1} and snapshots of the simulation run for 25 rescuers (Strategy 1) and 25 searchers are presented in Fig.  \ref{fig:snapShots}. \par

\par

Figure \ref{pic:TimeSeries1}, shows the time series graph of rate of retrieval of targets, for different agent combinations from Scenario 3 (constant populations) with rescuers following type Strategy 1. Note that the worst performing combinations were of those with the least number of rescuers. This directly correlates to the fact that lesser rescuers means lesser retrieval rates. 

\begin{figure}[h]
\centering
\vspace{-4mm}
\includegraphics[width=8.5cm]{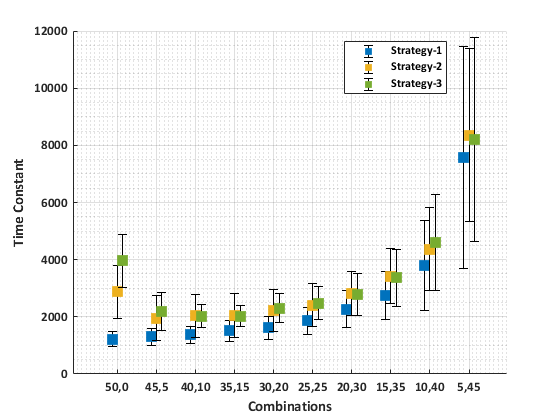}
\setlength{\belowcaptionskip}{-15pt}
\caption{Average time constants for different strategies in scenario 3, for (Rescuer, Searcher) combinations.}
\label{pic:TimeConst3}
\end{figure}

Further, for a better comparison of performance, we consider the average time constant across all the trials in each scenario and strategy combinations. Here, we define a time constant as the amount of time taken to retrieve 63\% of the targets ($158)$.The mean and standard deviation graphs of the time constants values computed for different strategy and scenario combinations are presented in Figs.  \ref{pic:TimeConst1}, \ref{pic:TimeConst2}, and \ref{pic:TimeConst3}.

\subsection{Time Constant Analysis}
\begin{figure}[h]

\centering
\includegraphics[width=8.5cm]{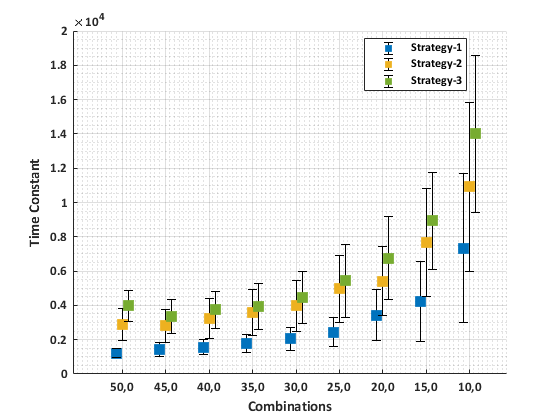}
\setlength{\belowcaptionskip}{-15pt}
\caption{Average time constants for different strategies in scenario 1, for (rescuer, Searcher) combinations.}
\label{pic:TimeConst1}

\end{figure}

The time constant results for Scenario 1 across all the three strategies involving homogeneous groups with no searchers, is shown in Fig.  \ref{pic:TimeConst1}. It can be noted that with the decreasing number of rescuers, the time constant increased, as there are more targets and lesser workers. And further, of all the strategies, Strategy 3 rescuers performed the worst due to their lack of ability to communicate with other rescuers, and locate long range targets by themselves. This is followed by the Strategy 2 rescuers' performance, where the rescuers have all the sensors enabled, however, lack the ability to communicate with other rescuers.

\begin{figure}[h]
\centering
\includegraphics[width=8.5cm]{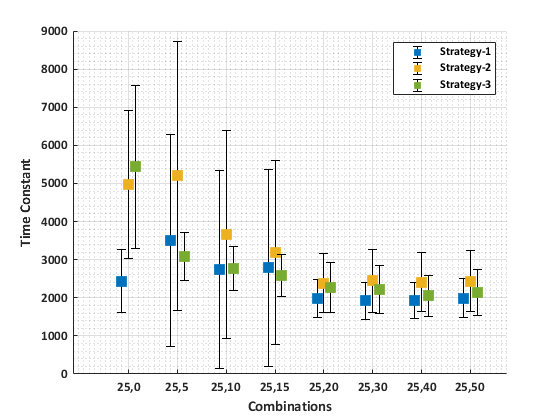}
\setlength{\belowcaptionskip}{-10pt}
\caption{Average time constants for different strategies in scenario 2, for (rescuer, Searcher) combinations.}
\label{pic:TimeConst2}
\end{figure}
Further, in Scenario 2 (ref. Fig.  \ref{pic:TimeConst2}), with a constant rescuer population and varying searchers, the addition of searchers to a homogeneous system showed some fluctuations in the beginning. However, the values settled at significantly lower average time constant in the end across all the strategies. Of all, strategy 3 showed the highest change as the rescuers were solely dependent on the searchers for the target locations as they were blind to long range target detection. Also, the time constant values not changing with additional rescuers after a certain threshold, indicates that the system reached saturation and this also correlates to the change in  heterogeneity measure in Fig.~\ref{pic:heteroMeasurea}.

In Scenario 3, we kept the total population constant at 50 and varied the ratio of rescuer to searcher population. This graph best explains the importance of heterogeneity in a multi-agent system. The first combination in the graph shows a homogeneous configuration, with only rescuers in the group. In Strategies 2 and 3, the time constant constantly dropped until 15 searchers and steadily increased there after. It can be noted that the heterogeneity measure for this scenario increases towards the center of the graph while it is more homogeneous towards the ends as also observed in Fig.  \ref{pic:heteroMeasureb}. This means a dip at the center is an indicator of a better performing heterogeneous system. Also, in some instances, Strategy 3 showed a lower time constant compared to Strategy 2, though the rescuers were blind, clearly demonstrating the benefits of heterogeneity. However, this does not hold good for the Strategy 1 type rescuers.\par

In Strategy 1, the rescuers are more capable than searchers compared to the rescuers in the other two strategies. Hence, any drop in the rescuers count was counter acting the heterogeneity benefits as the overall capability of the group was hampered.
This also indicates the distribution of functionalities as a major factor of mission efficiency. A team with a no functional overlap showed a better performance compared to larger functional overlap between the agents i.e., rescuers with more searcher capabilities. 

\subsection{Cost and Efficiency}
We further analyzed the efficiency of the system by associating a cost to each of the robots, for scenario 3 and all strategies. 
This cost is proportional to the functional abilities of the robots. From the definition of the rescuer and searchers, it can be clearly understood that a searcher costs lower than a rescuer, as a rescuer also has to perform the target retrieval task , which involves a pickup and delivery process. We combine this cost factor with the time constant from the previous graphs and we define efficiency as
\begin{equation}
    Efficiency=\frac{1}{\tau(c.n_r+n_s)} ,
\end{equation}
where $\tau$ is average time constant, $c$ is a cost factor, which was varied across strategies for our current study.

\begin{figure}[t]
\centering
\includegraphics[width=9cm]{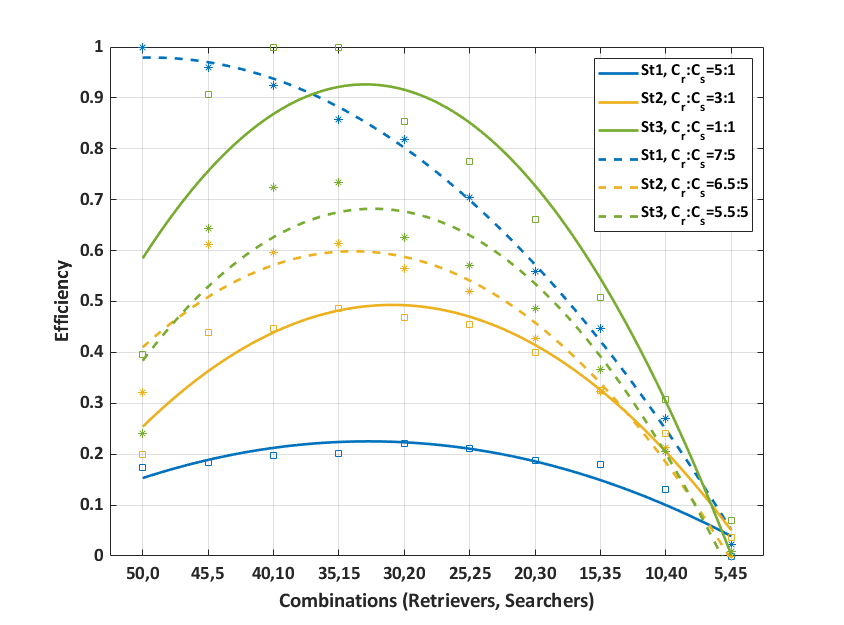}
\setlength{\belowcaptionskip}{-10pt}
\setlength{\abovecaptionskip}{-1pt}
\caption{Efficiency estimate of scenario 3}
\label{pic:costGraphA}
\vspace{-4mm}
\end{figure}

For the analysis presented in Fig. \ref{pic:costGraphA}, the cost ratios of rescuers ($C_r$) searchers ($C_s$) were chosen as 5:1, 7:5 for strategy 1, 3:1, 6.5:5 for strategy 2, and 1:1, 6.5:5 for strategy 3. 
These values were chosen proportional to the total inter-species distance computed in section \ref{sec:measure}, also expecting the searcher robots cost to be only a fraction of rescuer.And, in Strategy 3, we made rescuers cost closer to searchers as the inter-species distances are comparable. 

Though strategy 1 showed higher rate of retrieval as observed in the time constant graph (Fig. \ref{pic:TimeConst3}), the efficiency values were significantly lower when a cost was associated with the composition. Further, across the strategies, when the rescuer abilities were diminished along with the costs, the efficiency increased significantly. Also, in all the strategies, the peak efficiency was close towards the combinations with high heterogeneity measure.

\par

\section{Conclusions}
In this study, we questioned whether heterogeneity is beneficial in a heterogeneous multi-robot system. We modeled an search and rescue  problem for analyzing this influence. Our simulations and analysis indicate faster retrieval times proportional to the growth in heterogeneity measures in a searcher-rescuer team of robots. This supports our hypothesis that heterogeneity in a multi-robot system  is beneficial in general for enhancing system-level performance and also reducing global costs. However, there are exceptions. For example, in our Strategy 1, heterogeneity had a negative effect, where the high-performing rescuers reduces the searchers ability making the overall system less efficient when searchers replaces rescuers in Scenario 3. 

In other words, distribution of capabilities can play a significant role in enhancing performance through heterogeneous agent partnerships, a phenomenon demonstrated by symbiotic relationships between species in nature \cite{leung2008parasitism}. Further, our analysis study showed that robots of limited capabilities combined with other heterogeneous types can enhance the overall performance both in terms of cost and performance. This points to a new challenge of selection of right heterogeneity in functionalities across robots for maximising group efficiency, which can be treated as a multi-constraint optimization problem.


\bibliographystyle{IEEEtran}
\bibliography{references,newrefs}

\begin{thebibliography}{10}
\providecommand{\url}[1]{#1}
\csname url@samestyle\endcsname
\providecommand{\newblock}{\relax}
\providecommand{\bibinfo}[2]{#2}
\providecommand{\BIBentrySTDinterwordspacing}{\spaceskip=0pt\relax}
\providecommand{\BIBentryALTinterwordstretchfactor}{4}
\providecommand{\BIBentryALTinterwordspacing}{\spaceskip=\fontdimen2\font plus
\BIBentryALTinterwordstretchfactor\fontdimen3\font minus
  \fontdimen4\font\relax}
\providecommand{\BIBforeignlanguage}[2]{{%
\expandafter\ifx\csname l@#1\endcsname\relax
\typeout{** WARNING: IEEEtran.bst: No hyphenation pattern has been}%
\typeout{** loaded for the language `#1'. Using the pattern for}%
\typeout{** the default language instead.}%
\else
\language=\csname l@#1\endcsname
\fi
#2}}
\providecommand{\BIBdecl}{\relax}
\BIBdecl

\bibitem{cubber2017introduction}
G.~D. Cubber, D.~Doroftei, K.~Rudin, K.~Berns, A.~Matos, D.~Serrano,
  J.~Sanchez, S.~Govindaraj, J.~Bedkowski, R.~Roda \emph{et~al.},
  ``Introduction to the use of robotic tools for search and rescue,'' 2017.

\bibitem{erdelj2017help}
M.~Erdelj, E.~Natalizio, K.~R. Chowdhury, and I.~F. Akyildiz, ``Help from the
  sky: Leveraging uavs for disaster management,'' \emph{IEEE Pervasive
  Computing}, vol.~16, no.~1, pp. 24--32, 2017.

\bibitem{liu2016multirobot}
Y.~Liu and G.~Nejat, ``Multirobot cooperative learning for semiautonomous
  control in urban search and rescue applications,'' \emph{Journal of Field
  Robotics}, vol.~33, no.~4, pp. 512--536, 2016.

\bibitem{polka2017use}
M.~P{\'o}{\l}ka, S.~Ptak, and {\L}.~Kuziora, ``The use of uav's for search and
  rescue operations,'' \emph{Procedia engineering}, vol. 192, pp. 748--752,
  2017.

\bibitem{yang2020needs}
Q.~Yang and R.~Parasuraman, ``Needs-driven heterogeneous multi-robot
  cooperation in rescue missions,'' \emph{arXiv preprint arXiv:2009.00288},
  2020.

\bibitem{10.2307/3036036}
\BIBentryALTinterwordspacing
J.~L. Bronstein, ``Our current understanding of mutualism,'' \emph{The
  Quarterly Review of Biology}, vol.~69, no.~1, pp. 31--51, 1994. [Online].
  Available: \url{http://www.jstor.org/stable/3036036}
\BIBentrySTDinterwordspacing

\bibitem{Thompson372}
\BIBentryALTinterwordspacing
J.~N. Thompson, ``Mutualistic webs of species,'' \emph{Science}, vol. 312, no.
  5772, pp. 372--373, 2006. [Online]. Available:
  \url{https://science.sciencemag.org/content/312/5772/372}
\BIBentrySTDinterwordspacing

\bibitem{dales1957interrelations}
R.~P. Dales, ``Interrelations of organisms. a. commensalism,'' \emph{Treatise
  on marine ecology and paleoecology}, vol.~1, pp. 391--412, 1957.

\bibitem{wobeser2008parasitism}
G.~A. Wobeser, ``Parasitism: costs and effects,'' pp. 3--9, 2008.

\bibitem{twu2014measure}
P.~Twu, Y.~Mostofi, and M.~Egerstedt, ``A measure of heterogeneity in
  multi-agent systems,'' in \emph{2014 American Control Conference}.\hskip 1em
  plus 0.5em minus 0.4em\relax IEEE, 2014, pp. 3972--3977.

\bibitem{rizk2019cooperative}
Y.~Rizk, M.~Awad, and E.~W. Tunstel, ``Cooperative heterogeneous multi-robot
  systems: A survey,'' \emph{ACM Computing Surveys (CSUR)}, vol.~52, no.~2,
  p.~29, 2019.

\bibitem{pippin2011bayesian}
C.~E. Pippin and H.~Christensen, ``A bayesian formulation for auction-based
  task allocation in heterogeneous multi-agent teams,'' in \emph{Ground/Air
  Multisensor Interoperability, Integration, and Networking for Persistent ISR
  II}, vol. 8047.\hskip 1em plus 0.5em minus 0.4em\relax International Society
  for Optics and Photonics, 2011, p. 804710.

\bibitem{7440563}
M.~{Erdelj} and E.~{Natalizio}, ``Uav-assisted disaster management:
  Applications and open issues,'' in \emph{2016 International Conference on
  Computing, Networking and Communications (ICNC)}, Feb 2016, pp. 1--5.

\bibitem{dadvar2020multiagent}
M.~Dadvar, S.~Moazami, H.~R. Myler, and H.~Zargarzadeh, ``Multiagent task
  allocation in complementary teams: a hunter-and-gatherer approach,''
  \emph{Complexity}, vol. 2020, 2020.

\bibitem{darintsev2019methods}
O.~Darintsev, B.~Yudintsev, A.~Y. Alekseev, D.~Bogdanov, and A.~Migranov,
  ``Methods of a heterogeneous multi-agent robotic system group control,''
  \emph{Procedia Computer Science}, vol. 150, pp. 687--694, 2019.

\bibitem{hunt2014consensus}
S.~Hunt, Q.~Meng, C.~Hinde, and T.~Huang, ``A consensus-based grouping
  algorithm for multi-agent cooperative task allocation with complex
  requirements,'' \emph{Cognitive computation}, vol.~6, no.~3, pp. 338--350,
  2014.

\bibitem{liu2013robotic}
Y.~Liu and G.~Nejat, ``Robotic urban search and rescue: A survey from the
  control perspective,'' \emph{Journal of Intelligent \& Robotic Systems},
  vol.~72, no.~2, pp. 147--165, 2013.

\bibitem{colledanchise2018behavior}
M.~Colledanchise and P.~{\"O}gren, \emph{Behavior Trees in Robotics and Al: An
  Introduction}.\hskip 1em plus 0.5em minus 0.4em\relax CRC Press, 2018.

\bibitem{colledanchise2018learning}
M.~Colledanchise, R.~Parasuraman, and P.~{\"O}gren, ``Learning of behavior
  trees for autonomous agents,'' \emph{IEEE Transactions on Games}, vol.~11,
  no.~2, pp. 183--189, 2018.

\bibitem{rao1982diversity}
C.~R. Rao, ``Diversity and dissimilarity coefficients: a unified approach,''
  \emph{Theoretical population biology}, vol.~21, no.~1, pp. 24--43, 1982.

\bibitem{leung2008parasitism}
T.~Leung and R.~Poulin, ``Parasitism, commensalism, and mutualism: exploring
  the many shades of symbioses,'' \emph{Vie et Milieu}, vol.~58, no.~2, p. 107,
  2008.

\end{thebibliography}

\end{document}